\documentclass[journal]{IEEEtran}

\usepackage[noend]{algpseudocode}

\usepackage{algorithmicx,algorithm}

\usepackage{amsmath}
\usepackage{graphicx}
\usepackage{epstopdf}

\usepackage{amsthm,amsmath,amssymb}
\usepackage{amssymb}
\usepackage{mathrsfs}

\usepackage{dutchcal}
\usepackage{capt-of}
\usepackage{amsmath}

\usepackage{amssymb}

\ifCLASSINFOpdf
\else
\fi
\begin{document}
%
\title{Hypergraph Convolutional Network based Weakly Supervised Point Cloud Semantic Segmentation with Scene-Level Annotations}
%
%
%

\author{Zhuheng~Lu,
	Peng~Zhang,
	Yuewei~Dai,
	Weiqing~Li,
	and~Zhiyong~Su
	
	\thanks{Z. Lu, P. Zhang, and Z. Su are with the Visual Computing Group, School of Automation, Nanjing University of Science and Technology, Nanjing, Jiangsu Province 210094, P.R. China (e-mail: lzharsenal@163.com, zhangpeng@njust.edu.cn, su@njust.edu.cn)} 
	
	\thanks{Y. Dai is with the School of Electronic and Information Engineering, Nanjing University of Information Science and Technology, Nanjing, Jiangsu Province 210044, P.R. China (e-mail: dywjust@163.com).}
	
	\thanks{W. Li is with the School of Computer Science and Engineering, Nanjing University of Science and Technology, Nanjing, Jiangsu Province 210094, P.R. China (e-mail: li\_weiqing@njust.edu.cn).}
	
	\thanks{Manuscript received 00 00, 0000; revised 00 00, 0000. (Corresponding author: Zhiyong Su.)}}

%
%

\markboth{Journal of \LaTeX\ Class Files,~Vol.~14, No.~8, August~2015}%
{Shell \MakeLowercase{\textit{et al.}}: Bare Demo of IEEEtran.cls for IEEE Journals}
%



\maketitle

\begin{abstract}
Point cloud segmentation with scene-level annotations is a promising but challenging task.
Currently, the most popular way is to employ the class activation map (CAM) to locate discriminative regions and then generate point-level pseudo labels from scene-level annotations.
However, these methods always suffer from the point imbalance among categories, as well as the sparse and incomplete supervision from CAM.
In this paper, we propose a novel weighted hypergraph convolutional network-based method, called WHCN, to confront the challenges of learning point-wise labels from scene-level annotations. 
Firstly, in order to simultaneously overcome the point imbalance among different categories and reduce the model complexity, superpoints of a training point cloud are generated by exploiting the geometrically homogeneous partition.
Then, a hypergraph is constructed based on the high-confidence superpoint-level seeds which are converted from scene-level annotations.
Secondly, the WHCN takes the hypergraph as input and learns to predict high-precision point-level pseudo labels by label propagation.  
Besides the backbone network consisting of spectral hypergraph convolution blocks, a hyperedge attention module is learned to adjust the weights of hyperedges in the WHCN.
Finally, a segmentation network is trained by these pseudo point cloud labels. 
We comprehensively conduct experiments on the ScanNet and S3DIS segmentation datasets. 
Experimental results demonstrate that the proposed WHCN is effective to predict the point labels with scene annotations, and yields state-of-the-art results in the community.
The source code is available at http://zhiyongsu.github.io/Project/WHCN.html.
\end{abstract}


%
\IEEEpeerreviewmaketitle

\section{Introduction}\label{sec:introduction}
%
%
%
%
\IEEEPARstart{P}{oint} cloud segmentation is a fundamental technique in scene understanding. 
Benefiting from recent advances of deep learning methods, point cloud semantic segmentation has achieved remarkable progress \cite{qi2017pointnet,qi2017pointnet++,li2018pointcnn,wang2019dynamic,shuai2021backward,cheng2021net}. 
However, obtaining precise point-wise annotations for semantic segmentation tasks is quite exhausting and costly. 
Weakly supervised semantic segmentation is receiving great attention since it reduces the cost of annotations required to train models.
Some promising efforts are motivated to obtain point-level semantic prediction using weak annotations as supervision, such as 3D bounding box \cite{choy20194d}, scribble \cite{shu2019scribble}, point \cite{xu2020weakly}, and scene-level annotations \cite{wei2020multi}. 
Among these weak supervisions, scene-level annotation is one of the most economical and efficient supervision. 
In this context, each point cloud has its scene category labels (\emph{e.g.}, door, table, chair), which indicate the categories of objects in the point cloud.
Therefore, the key challenge of this problem lies in how to accurately estimate the pseudo object labels using the given scene-level supervision.

However, there are very a few works that employ scene-level supervision to segment point clouds. 
The typical methods leverage class activation map (CAM) to locate discriminative regions, and then generate point-level pseudo labels to train the segmentation network \cite{wei2020multi,ren20213d}. 
The main challenges of scene-level supervised semantic segmentation using CAM lies in two aspects.
The first is the point imbalance among categories. Taking the indoor scene for example, some categories (\emph{e.g.}, floor, wall) have a dominant number of points. 
Therefore, the imbalance is not conducive for the classification network to locate confident regions of small objects.
The second is that the activation maps obtained from the classification network are sparse and incomplete.
To solve the above problems, MPRM \cite{wei2020multi} attempts to use scene-level and subcloud-level labels for the weakly supervised segmentation task. 
However, there are some drawbacks that limit its performance. MPRM needs to generate extra subclouds and manually label these subclouds to deal with category imbalance issue. 
Meanwhile, the split of the subclouds also inevitably loses the global structural information.
WyPR \cite{ren20213d} tries to learn a joint 3D segmentation and detection
model using scene-level labels. 
They convert the per-point logits into a scene-level prediction using average pooling and a sigmoid normalization.
However, the segmentation mask generated by WyPR is incomplete since it focuses on the points with high scores.
All in all, weakly supervised semantic segmentation based on scene annotations is extremely challenging since it is difficult to obtain fine-grained point-level annotations from scene-level annotations.

In this paper, to address the above two critical issues, we propose a weighted hypergraph convolutional network-based method (WHCN) for weakly supervised semantic segmentation:

(1) For the first issue, we generate superpoints to balance the point numbers among different categories, where each point cloud is divided into primitive patches. 
Besides, the superpoint generation module also reduces the complexity of the original hypergraph model.
Then, we generate superpoint-level seed labels using scene-level annotations. Meanwhile, the selected confident seed labels will be converted to corresponding vertex labels.

\begin{figure}[!t]
	\centering
	\includegraphics[width=0.4\textwidth]{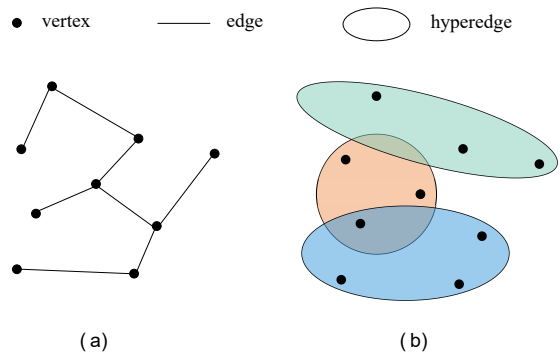}\\
	\caption{The difference between (a) a graph and a (b) hypergraph. In a graph, each edge, represented by a line, only connects two vertices. In a hypergraph, each edge, represented by a colored ellipse, connects more than two vertices.}
	\label{fig:hypergraph}
\end{figure}

(2) For the second issue, we design a novel weighted hypergraph convolutional network, called WHCN, to generate accurate point-level pseudo labels for weakly supervised semantic segmentation. 
Our approach builds upon the hypergraph representation of a point cloud, which is a high-dimensional graph model. Fig \ref{fig:hypergraph} shows the difference between a graph and a hypergraph.
It can be seen that a hypergraph can capture the between-vertex relationships that are beyond pairwise. 
To capture high-order semantic relations automatically, the WHCN is trained to predict pseudo labels for all vertices in the hypergraph. 
Then, the final point-level pseudo labels are generated by mapping superpoint labels to corresponding points in the superpoint.

To validate the effectiveness of our WHCN, we perform extensive experiments on ScanNet and S3DIS datasets. According to the experimental results, we achieve new state-of-the-art performances among all these tasks with scene-level annotations.

The main contribution of this work are as follows: 

\begin{itemize}
	\item We propose a framework that effectively takes the advantage of Hypergraph Neural Network to learn high-order semantic relations for scene-level supervised point cloud semantic segmentation. 
	
	\item We utilize the superpoint generation module to balance the point numbers among different categories by segmenting locally homogeneous and coherent regions, which also can reduce the model complexity.   
	
	\item We propose a label propagation module, WHCN, to discover more objects in non-salient regions through feature aggregation based on related vertices and hyperedges.

\end{itemize}

The rest of the paper is organized as follows. Section \ref{sec:related work} briefly reviews the related work on semantic segmentation for point cloud and graph neural networks. In Section \ref{sec.overview}, an overview  of the framework is provided. Section \ref{sec.construction} describes the details of hypergraph construction. In Section \ref{sec.pseudo}, the details of pseudo label generation are introduced. After that, experimental results are presented in Section \ref{sec.experiment}. Finally, conclusions and recommendations for future research are given in Section \ref{sec.conclusion}.

\section{Related Work}\label{sec:related work}
In this section, we first introduce related work on semantic segmentation for point cloud. Then, we review a number of previous works on weakly supervised semantic segmentation task both in 3D point clouds. Finally, it is essential for us to review the Graph Neural Networks since our model is based on the hypergraph.

\subsection{Supervised Semantic Segmentation for Point Cloud}

In general, existing supervised approaches for point cloud semantic segmentation can be roughly divided into point-based, projection-based, and voxel-based methods.
Point-based methods directly take raw point clouds as input and extract semantic information by point convolutions. 
PointNet \cite{qi2017pointnet} and PointNet++ \cite{qi2017pointnet++} are two pioneering works and provide the prominent advances of deep neural networks.
A series of other works \cite{li2018pointcnn,wang2019dynamic,thomas2019kpconv,boulch2020convpoint,wu2019pointconv,hu2020randla,ye20183d} also explore different strategies for leveraging local structure learned from point clouds.
These frameworks tend to be computationally efficient and have the potential to preserve semantics for every single 3D point.
Projection-based methods \cite{boulch2017unstructured,lyu2020learning,milioto2019rangenet++,xu2020squeezesegv3,lawin2017deep} project point clouds to 2D planes and use multi-view or spherical images as representations. 
Taking advantage of the power of CNNs in image processing, projection-based methods achieve reasonable 3D segmentation performance. However, critical geometric information is likely to be lost in the projection step.
Voxel-based methods \cite{wu20153d,riegler2017octnet,su2018splatnet,graham20183d,le2018pointgrid,meng2019vv,zhang2020polarnet} convert a point cloud into a volumetric representation, and take the regular voxel-grids as input.
Their weaknesses are quantization artifacts, inefficient usage of 3D voxels, and low spatial resolutions due to a large memory requirement.


\begin{figure*}[!t]
	\centering
	\includegraphics[width=0.95\textwidth]{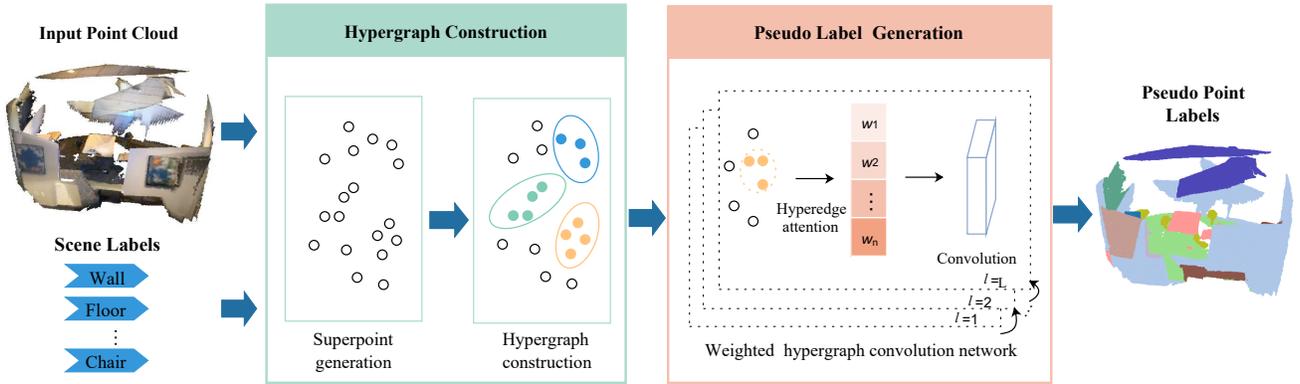}\\
	\caption{Overview of our proposed WHCN framework. First, superpoints are generated from the input point clouds. Meanwhile, scene-level annotations are transferred to superpoint-level labels. The hypergraph is constructed based on the generated superpoints. Then, pseudo labels are learned from the proposed weighted hypergraph convolution network. }
	\label{fig:framework}
\end{figure*}

\subsection{Weakly Supervised Semantic Segmentation for Point Cloud}
To avoid the exhausting annotation, weakly supervised semantic segmentation attempts to learn a segmentation network with weaker annotations. 
Compared to the weakly supervised image segmentation, weakly supervised semantic segmentation of point clouds is relatively unexplored.
Xu et al. \cite{xu2020weakly} aimed to learn a point cloud segmentation model with only partially labeled points. They labeled only 10\% of the points in the training stage. 
Shu et al. \cite{shu2019scribble} used sparse scribble-based labels for 3D shapes segmentation. The labeling information was automatically propagated to the unlabeled parts of the training 3D shapes and given as a part of the optimization results.
Wei et al. \cite{wei2020multi} proposed MPRM that uses both scene-level and subcloud-level annotation to generate point-level pseudo labels. They uniformly generated seed points in the space and take all the neighboring points within a specific radius to form a subcloud. Then, they used these pseudo labels to train a point cloud segmentation network in a fully supervised manner.
Ren et al. \cite{ren20213d} designed WyPR to jointly learn semantic segmentation and object detection for point clouds from only scene-level category labels. 
However, the region masks obtained from two scene-level methods were sparse and incomplete since they focused on locating the most discriminative part of objects. 
Inspired by the weakly supervised image segmentation, we propose a label propagation module to generate accurate pseudo labels.

\subsection{Graph Neural Network}

As a classic kind of data structure, graphs \cite{gori2005new,scarselli2008graph,pedronette2018unsupervised} can express not only the intrinsic entities with nodes but also the complicated relationships between the corresponding entities with their edges, which is commonly adopted in social networks, protein-interaction networks, knowledge graphs, etc.

Recently, graphs and graph signal processing have found applications in modeling point clouds, since their superiority in representation learning compared with traditional neural networks. Some works created structure from point clouds by representing it with multiple 2D views \cite{su2015multi,boulch2017unstructured,guerry2017snapnet}. 
Meanwhile, others \cite{2017ScanNet,riegler2017octnet,xu2020grid} generated graph models by representing point clouds with voxelization.
More recent works focused on directly representing unordered point clouds with graph structure. 
Wang et al. \cite{wang2019dynamic} proposed a dynamic edge convolution algorithm for semantic segmentation of point clouds. The EdgeConv dynamically computed node adjacency at each graph layer using the distance between point features. 
Li et al. \cite{li2019deepgcns} aimed to train very deep GCNs in the task of point cloud semantic segmentation. They adapted residual connections, dense connections, and dilated convolutions from CNN models to GCN architectures. 

As stated above, most existing variants of GNN assume pairwise relationships between objects, while our work introduces the hypergraph structure to model high-order relations among data.
As a generalization of a graph, hypergraph learning was first introduced by Zhou et al. \cite{zhou2007learning}, to conduct label propagation on hypergraph structure for classification and clustering.
It has been further applied in other tasks such as object retrieval and recognition \cite{gao20123,zhang2018cross}, object classification \cite{zhang2018inductive,wei2015combinative,nong2021hypergraph} and segmentation \cite{huang2009video,2021lv4D}.
Recently, some hypergraph neural networks have been proposed for deep learning on hypergraphs. 
They can be divided into two categories according to the way they deal with a hypergraph. 
One branch of algorithms is to map hypergraph into a general graph, to which graph convolution approaches can be applied.
These methods \cite{yadati2019hypergcn,bandyopadhyay2020line} increased the difficulty of transformation and cannot retain high-order information well.
Another kind of methods is to design the convolution operator directly on a hypergraph. Feng et al. \cite{2019hypergraph} proposed a Hypergraph neural networks (HGNN) framework, in which hyperedge convolution was designed by using the hypergraph Laplacian, so that the output vertex feature was obtained by aggregating their related hyperedge feature. Jiang et al. \cite{jiang2019dynamic} added a dynamic hypergraph construction module to hypergraph convolution, and designed vertex convolution and hyperedge convolution to aggregate features of vertices and hyperedges, respectively.

\section{Overview}\label{sec.overview}
In this paper, we focus on the task of weakly supervised semantic segmentation with scene-level labels. Our framework mainly consists of two parts: hypergraph construction and pseudo label generation, as illustrated in Fig.\ref{fig:framework}.

Hypergraph construction converts the input point clouds to hypergraphs. We first generate superpoints using the geometrically homogeneous partition strategy to balance the point numbers among different categories.
Specifically, for an input point cloud, we use a global energy model to generate superpoints. 
Therefore, the point-wise segmentation problem is reformulated as a superpoint-wise segmentation problem. 
Then, we use the class activation map \cite{zhou2016learning} to convert scene-level labels to superpoint-level labels.
In addition, to guarantee the accuracy of supervision, we only choose a limited number of confident seed labels as supervision. 
In the hyperedge construction, the generated superpoints constitute the vertex set, and
each hyperedge is formed by the labeled superpoint seeds of the corresponding category.

Pseudo label generation generates final point-level pseudo labels. 
WHCN presents an effective label propagation scheme that exploits relations among the vertices with same properties. 
Firstly, we propose a hyperedge attention module to obtain affinity for each vertex set associated with a hyperedge and better provide propagation information.
Then, we introduce an essential operator for spectral hypergraph convolution.
Thus, the feature aggregation between property-wise vertices can ensure accurate propagation.
Finally, we train a point cloud segmentation network using the generated point-level pseudo labels.

\section{Hypergraph Construction}\label{sec.construction}
In this section, we elaborate on the phase of hypergraph construction for the weakly supervised semantic segmentation task. Firstly, given a point cloud, we generate superpoints using a global energy model. Then, we leverage the CAM to generate superpoint-level seeds. Finally, we build the hypergraph to effectively represent a point cloud.

\subsection{Superpoints Generation}\label{sec.SG}
First of all, the superpoint method \cite{2018Large} is applied to produce the hypergraph vertices. Similar to the superpixel methods for image segmentation, superpoints are suitable as nodes of the graph structure since they can produce a locally homogeneous and coherent region. 

To generate a set of superpoints $ \mathcal{S}=\{s_{1},s_{2},\dots,s_{N}\} $, firstly, we use a global energy model described in \cite{2017weakly,2018Large}. 
It is a strategy to compute piecewise constant functions on graphs. 
Let $ G=(V,E) $ be the graph representing the input point cloud, where each node in $ V $ represents a point, and each point is connected to its $K$ nearest neighbors. For each point, we compute its local geometric features $ f_{i}\in \mathbb{R}^{d}$, which consist of linearity, planarity, scattering, and verticality introduced by \cite{2017weakly}.
Then, the geometrically homogeneous partition can be defined as computing a piecewise constant approximation $g'$ of the signal $f\in\mathbb{R}^{d\times \vert V\vert} $. $g'$ can be obtained by minimizing the following energy: 
\begin{equation}\label{eq.1}
g'=\displaystyle{\arg \min_{g\in\mathbb{R}^{d\times \vert V\vert}}}\displaystyle\sum_{i\in \vert V\vert}\Vert g_{i}-f_{i} \Vert^{2}+\rho\displaystyle\sum_{(i,j)\in E}\delta(g_{i}-g_{j}\neq 0) 
\end{equation}
where $g_{i}\in\mathbb{R}^{d} $, $g\in\mathbb{R}^{d\times \vert V\vert} $ is the optimization variable, $ \delta(\cdot \neq 0) $ defines the function of $ \mathbb{R}^{d}\mapsto \{0,1\} $, which is equal to 0 if $g_{i}-g_{j}=0 $, while 1 everywhere else. The factor $ \rho  $ is the regularization strength and determines the number of the partitions.  

The optimization problem defined in Eq. (\ref{eq.1}) can be efficiently approximated with the greedy graph-cut based $ \mathcal{l}_{0} $-cut pursuit algorithm presented in \cite{2017cut}. 
$ \mathcal{l}_{0} $-cut pursuit proceeds in a top-down manner, by iteratively splitting the data, and has no prior on the number of regions. 
The solutions of Eq. (\ref{eq.1}) are referred as the set of superpoints $ \mathcal{S}=\{s_{1},s_{2},\dots,s_{N}\} $. 
Then, the superpoints generation step reformulates the point-wise segmentation problem as a superpoint-wise segmentation problem, where the superpoint labels correspond to the final pseudo labels of the associated points.
Finally, we employ the efficient PointNet \cite{qi2017pointnet}, following \cite{2018Large}, to independently infer a descriptor for each superpoint by embedding it into a fixed dimensional feature vector $ f(s) $.

\begin{figure}[!t]
	\centering
	\includegraphics[width=0.47\textwidth]{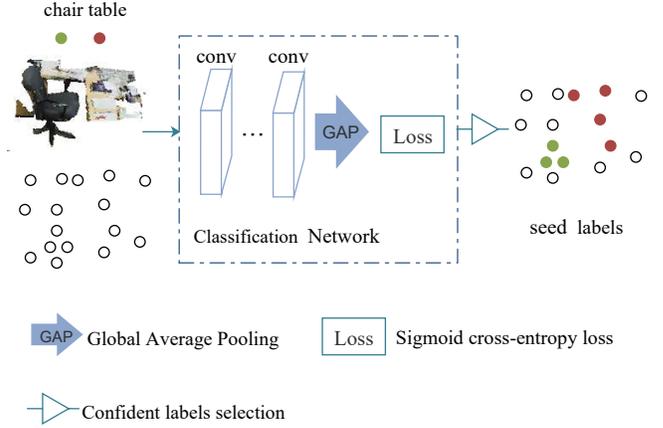}\\
	\caption{Overview of the superpoint seed label generation. We input the generated superpoints and the corresponding scene labels. Then, we train a classification network to generate the class activation maps. Meanwhile, we select confident superpoint-level seed labels as the vertex labels. The labels of white vertices are unknown.}
	\label{fig:seed}
\end{figure}

\subsection{Superpoint Seed Label Generation}

To generate superpoint-level labels from scene-level annotations, a CAM-based method is used to generate the initial seed labels for propagation as shown in Fig. \ref{fig:seed}. 
Specifically, we employ a classification network, such as PointNet++ \cite{qi2017pointnet++}, to generate class attention maps. 
We input the superpoint set $ \mathcal{S}=\{s_{1},s_{2},\dots,s_{N}\} $ and the corresponding scene labels (\emph{e.g.}, table, chair) to the classification network. During training, we take the superpoint prediction using the global average pooling layer and calculate a sigmoid cross-entropy loss with scene labels.
Then, for the category $c$, the class activation map $M_{c}$ for the superpoint $s_{k}\in \mathcal{S}$ can be obtained by:
\begin{equation}\label{eq.seed}
M_{c}(s_{k})= {\mathrm{w}_{c}}^{\intercal}\cdot f(s_{k})
\end{equation}
where $ \mathrm{w}_{c} $ is the classifier weight for category $ c $.

$M_{c}(s_{k})$ indicates the importance of the activation at superpoint $s_{k}$ corresponding to the category $ c $. Hence, the pseudo label of superpoint $s_{k}$ is calculated by $argmax(M(s_{k}))$. We only select those highly confident superpoint-level seed labels as the final seed labels with an adaptive threshold to remove the noisy labels. Specifically, we select the top 40\% confident point labels as seed labels from the Eq.(\ref{eq.seed}) following \cite{zhang2021affinity}.

\subsection{Hypergraph Construction} \label{sec.hyper}

\begin{figure*}[!t]
	\centering
	\includegraphics[width=0.95\textwidth]{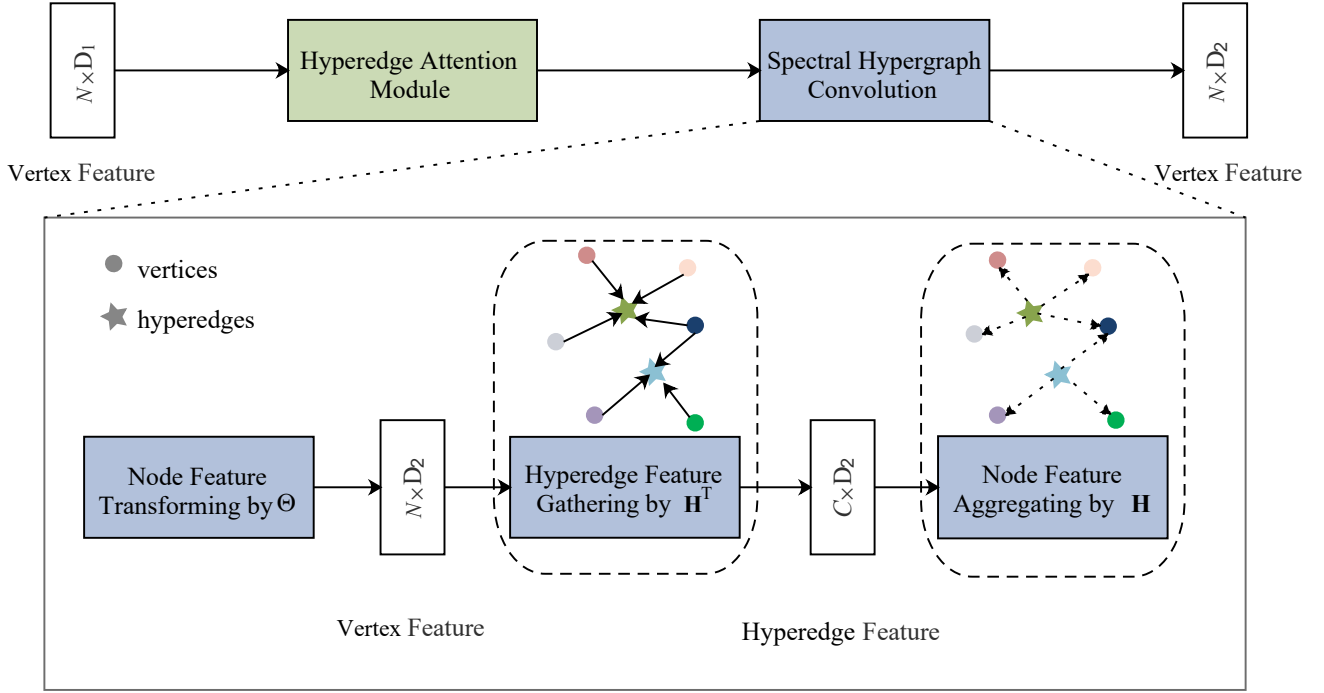}\\
	\caption{Illustration of the hypergraph convolution layer. The initial vertex feature is transformed by a learnable matrix  $ \mathbf{\Theta} $. Then, the new vertex feature on the hyperedges is gathered to obtain the hyperedge feature by $\mathbf{{H}}^{\intercal}$. Finally, the related hyperedge feature is associated to obtain the final vertex feature using the matrix $\mathbf{{H}}$.} 
	\label{fig:convolution}
\end{figure*}

We define a hypergraph $\mathcal{G}=(\mathcal{V},\mathcal{\varepsilon})$, where $ \mathcal{V} $ denotes the vertex set, and $ \mathcal{\varepsilon} $ denotes the hyperedge set. 
In the construction, the generated $N$ superpoints constitute the vertex set $\mathcal{V} $ of the hypergraph. Let $T$ be the number of labeled vertices. Then, the corresponding labeled and unlabeled vertex sets are $\{v_{1},\cdots,v_{T}\}$ and $\{v_{T+1},\cdots,v_{N}\}$, respectively. 
A hyperedge is formed by the generated labeled superpoint seeds of the corresponding category. 
In addition, each hyperedge $ e\in \mathcal{\varepsilon} $ is assigned a positive weight $ w(e) $, with all the weights stored in a diagonal matrix $ \mathbf{W}\in \mathbb{R}^{C\times C}  $. 
$C $ represents the number of categories. 
Further, a hypergraph can be represented by an incidence matrix $ \mathbf{H} \in \mathbb{R}^{N\times C}$ , which is defined as:

\begin{equation}
h(v,e)=\begin{cases}

1 ,& \text{if} \  v \in e; \\

0 ,& \text{if} \  v \notin  e.

\end{cases}
\end{equation}

Then, the degree of a vertex $ v\in \mathcal{V} $ is denoted as:

\begin{equation}
d(v)=\sum w(e)h(v,e)
\end{equation}
where $ w(e) $ is a positive weight assigned to the hyperedge. The degree of a hyperedge $ e\in \mathcal{\varepsilon} $ is denoted as:

\begin{equation}
b(e)=\sum h(v,e)
\end{equation}

\section{Pseudo Label Generation}\label{sec.pseudo}
In this paper, we consider the segmentation problem as vertex label propagation on the hypergraph. 
Therefore, we propose a WHCN to propagate the labels of a small number of vertices to the unlabeled vertices using the available labels as supervision.
The WHCN is based on spectral hypergraph convolutional neural networks, as introduced in \cite{2019hypergraph,bai2021hypergraph}.
With the hyperedge attention module, WHCN learns dynamic hyperedge weights to ensure accurate propagation.

\subsection{WHCN Architecture}

The overall architecture of WHCN is a label propagation framework. 
Given the labeled vertex subset $ \{v_{1},\cdots,v_{T}\} \in \mathcal{V}$, WHCN predicts the labels associated with the remaining unlabeled vertices. 
Given the input feature matrix $ \mathbf{X}$ and the incidence matrix $ \mathbf{H}$, our goal is to learn the propagation function $F$:
\begin{equation}
F(\mathbf{X},\mathbf{H})=\mathbf{X}_{out}
\end{equation}
where $\mathbf{X}_{out}$ and $\mathbf{X}$ represent the output and input feature matrix, respectively.

To capture high-order semantic relations automatically, we use the hypergraph structure to better aggregate the features.
The pipeline of the proposed hypergraph convolution and hyperedge attention convolutional network is illustrated in Fig. \ref{fig:convolution}.
We train a spectral hypergraph convolution-based model $F$ with a supervised loss for all labeled vertices. 
To obtain the affinity between associated vertices and better propagate labels, we propose a hyperedge attention module.
Then, the weighted hypergraph convolution layer can perform vertex-edge-vertex transform.
Specifically, the process of the hypergraph convolution layer is as follows. Firstly, the initial vertex feature is transformed by a learnable filter matrix  $ \mathbf{\Theta} $. Then, the new vertex feature on the hyperedges is gathered to obtain the hyperedge features by the multiplication of $\mathbf{{H}}^{\intercal}$. Finally, the related hyperedge feature is associated to obtain the final output vertex feature by multiplying the matrix $\mathbf{{H}}$. 

\subsection{Hyperedge Attention Module}

The hyperedge weight describes the affinity attribute of each hyperedge \cite{agarwal2006higher}, which is the similarity measure of the vertex set associated with the hyperedge.
Therefore, the larger weight indicates a stronger relationship between the vertices in the hyperedge.
However, the existed works \cite{2019hypergraph,nong2021hypergraph,bai2021hypergraph} consider that all hyperedges in a hypergraph have identical weights. 
The identical weighted hypergraph can not accurately provide propagation information since it treats the hyperedges equally.
In this paper, we follow the rule that hyperedges with larger weights deserve more confidence in label propagation.

In this circumstance, the hyperedge attention module produces adaptive weights to better reveal the intrinsic relationship between vertices.
For each hyperedge $e$ and its $R$ associated vertices $\mathcal{v}=\{v_{1},\cdots,v_{R}\}$, inspired by \cite{veli2018graph}, the hyperedge weight $w(e)$ is defined as

\begin{equation}
w(e)= \frac{1}{R(R-1)}\displaystyle\sum_{\{v_{i},v_{j}\}\in\mathcal{v}}\exp\left(-\frac{\sigma (sim(\mathbf{x}_{i}\mathbf{\Theta},\mathbf{x}_{j}\mathbf{\Theta}))}{\mu}\right)
\end{equation}
where $\mu$ is a positive parameter that controls the scaling of the hyperedge weight, $ \sigma(\cdot) $ is the LeakyReLU nonlinearity, and $sim(\cdot) $ is a similarity function that computes the pairwise similarity between two vertices defined as

\begin{equation}
sim(\mathbf{x}_{i},\mathbf{x}_{j})=\mathbf{a}^{\intercal}[\mathbf{x}_{i}\parallel \mathbf{x}_{j}]
\end{equation}
where $[,\parallel,]$ denotes concatenation operation, and $\mathbf{a}$ is a weight vector used to output a scalar similarity value. 

\subsection{Spectral Hypergraph Convolution}

The general objective function of transductive hypergraph learning can be formulated as a regularization framework 
\begin{equation}
\mathrm{arg}\ \underset{F}{\mathrm{min}}\{R_{\mathrm{emp}}(F)+\alpha\Omega(F)\}
\end{equation}
where $ R_{\mathrm{emp}}(F) $ denotes the supervised empirical loss, such as the least square loss or cross entropy loss, $ \Omega(F) $ is a regularizer on hypergraph, and $ \alpha \textgreater 0 $ is the regularization parameter to balance the empirical loss and the regularizer. 

To perform the spectral convolution on the hypergraph, we adopt the normalized hypergraph Laplacian introduced in \cite{zhou2007learning}, which can be defined as:
\begin{equation}
\Delta=\mathbf{I}-\mathbf{D}^{-1/2}\mathbf{HWB}^{-1}\mathbf{{H}}^{\intercal}\mathbf{D}^{-1/2}
\end{equation}
where $ \mathbf{D} $ and $ \mathbf{B} $ are degree matrices of the vertex and hyperedge in a hypergraph, respectively, $  \mathbf{I}  $ is the identity matrix, $\mathbf{W}$ is the diagonal hyperedge weight matrix as described in Section \ref{sec.hyper}.

The hypergraph Laplacian $ \Delta $ is a positive semi-definite matrix, which can be eigendecomposed as $ \Delta=\mathbf{U}\Lambda\mathbf{U}^{\intercal}  $, where $ \mathbf{U}=(\mathbf{u_{1}},\mathbf{u_{2}},\dots,\mathbf{u}_{N}) $ is eigen vectors matrix, and $ \Lambda=diag(\lambda_{1},\lambda_{1},\dots,\lambda_{N}) $ is a diagonal matrix containing corresponding non-negative eigenvalues. Then, the Fourier transform for a hypergraph signal $ \mathbf{x}\in\mathbb{R}^{N} $ is defined as $ \hat{\mathbf{x}}=\mathbf{U}^{\intercal}\mathbf{x} $.

Hence, the spectral hypergraph convolution is defined as 

\begin{equation}
\mathbf{g}\star\mathbf{x}=\mathbf{U}((\mathbf{U}^{\intercal}\mathbf{g})\odot(\mathbf{U}^{\intercal}\mathbf{x}))=\mathbf{U}g( \Lambda)\mathbf{U}^{\intercal}\mathbf{x}
\end{equation}
where $\odot$ is the element-wise Hadamard product, and $g( \Lambda)=\mathbf{diag}(g( \lambda_{1}),g( \lambda_{1}),\dots,g( \lambda_{N}))$ is a function of the Fourier coefficients. To reduce the computational cost, the truncated Chebyshev polynomial is used to parameterize the filter. Then, the parameterized convolution can be written as:

\begin{equation}
\mathbf{g}\star\mathbf{x}\approx(\displaystyle\sum_{q=0}^{Q}\theta_{q}T_{q}(\displaystyle\widetilde{\Delta}))\mathbf{x}
\end{equation}
where $T_{q}(\displaystyle\widetilde{\Delta})$ is the Chebyshev polynomial of $q$-th order with scaled Laplacian $\widetilde{\Delta}=(2/\lambda_{max})\Delta-\mathbf{I}$, $\theta_{q}$ is learnable parameter, and $\lambda_{max}$ denotes the largest eigenvalue in the hypergraph Laplacian.

According to the propagation rules of classical neural networks, each layer can be written as a nonlinear function.
Hence, the layer-wise propagation rules can be defined as the following formulation:
\begin{equation}\label{eq.propagation}
\mathbf{X}^{(l+1)}=\sigma(\mathbf{D}^{-1/2}\mathbf{HWB}^{-1}\mathbf{{H}}^{\intercal}\mathbf{D}^{-1/2}\mathbf{X}^{(l)}\mathbf{\Theta}^{(l)})
\end{equation}
where $ \mathbf{X}^{(l)}$ is the input signal of $l$-th layer, $ \sigma(\cdot) $ is the nonlinear activation function, and $ \mathbf{\Theta}^{(l)} $ is the learnable weight matrix between the $ l $-th and $(l+1)$-th layer. 

\begin{figure*}[!t]
	\centering
	\includegraphics[width=0.9\textwidth]{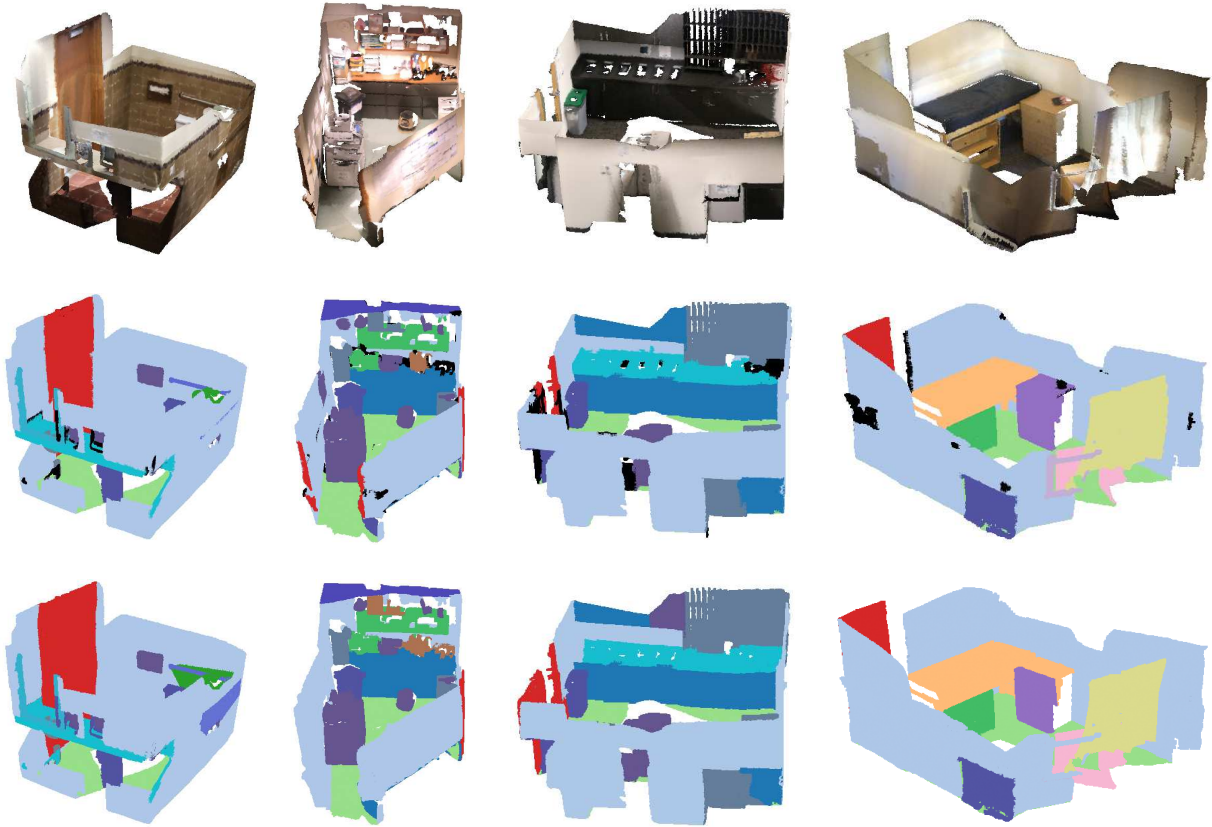}\\
	\caption{Qualitative results on the ScanNet dataset of our WHCN. The top row is the original input point clouds. The middle row is the ground truth. The bottom row shows the segmentation results with WHCN. The bottom row shows the segmentation results with WHCN. Note that the black points in the ground truth indicate unclassified points, which are ignored during evaluation.} 
	\label{fig:resultScanNet}
\end{figure*}

\subsection{Loss Function}

We train the model by minimizing the cross-entropy loss on the labeled vertices.
WHCN assigns a category label to each superpoint after training on the supervised loss $ \mathcal{L} $ for all vertices with labels

\begin{equation}
\mathcal{L}=-\displaystyle\sum_{i=1}^{T}\displaystyle\sum_{j=1}^{C}\mathbf{Y}_{ij}{\rm log}\ (p_{ij})
\end{equation}
where $\mathbf{Y}\in\mathbb{R}^{L\times C} $ denotes the ground truth label, $\mathbf{Y}_{ij}=1$ if the label of vertex $v_{i}$ is $j$, and $\mathbf{Y}_{ij}=0$ otherwise. $T $ and $C$ denote the number of labeled vertices and categories, respectively. $p_{ij} $ is the probability of the vertex $v_{i}$ belongs to category $j$.

\section{Experiments}\label{sec.experiment}

In this section, we evaluate the proposed WHCN on two public benchmark datasets for the semantic segmentation task. We first introduce the datasets and implementation details, and then present the experimental results.

\subsection{Datasets and Evaluation Metrics}

To demonstrate the effectiveness of our proposed method, we empirically evaluate WHCN on two large-scale datasets ScanNet \cite{2017ScanNet} and S3DIS \cite{armeni2017joint}. 

\textbf{ScanNet}. ScanNet is a large 3D dataset containing 1,513 point clouds of scans from 707 unique indoor scenes. ScanNet contains a variety of spaces such as offices, apartments, and bathrooms. The annotation of the point clouds corresponds to 20 semantic categories plus one for the unannotated space. We adopt the official train-val split, where there are 1205 training scenes and 312 validation scenes.

\textbf{S3DIS}. S3DIS is produced for indoor scene understanding and is widely used in the point cloud semantic segmentation task. S3DIS consists of 3D RGB point clouds of six floors from three different buildings split into individual rooms. Each room is scanned with RGBD sensors and is represented by a point cloud with coordinate information and RGB value. 
It contains 3D scans of 271 rooms with 13 categories, where area 5 is used for testing and the rest for training.  

\textbf{Evaluation metric}.
For all the experiments, we only adopt the scene-level category labels for training.
Standard mean intersection over union (mIoU) is applied as the evaluation criterion for the weakly supervised semantic segmentation.

\subsection{Implementation Details}
We implement the proposed framework using Pytorch. 
For the classification network, we adopt the PointNet++\cite{qi2017pointnet++} classification network as our backbone due to its remarkable robustness, efficiency, and simplicity. It is pretrained on ScanNet and S3DIS. 
Besides, we use the KPConv\cite{thomas2019kpconv} segmentation model as our final segmentation model for a fair comparison with \cite{wei2020multi}.
Following \cite{2018Large}, we set $\rho$=0.03 and select 512 superpoints.
In more detail, We build the network architecture with two hypergraph convolutional layers. 
The feature dimension of the hidden layer is set to 128.
We employ the ReLU as the nonlinear activation function, and use dropout to avoid overfitting with drop rate $p$=0.5. 
Then, the softmax function is used to generate predicted labels. $K$ in section \ref{sec.SG} is set as 10 followed by \cite{2017weakly,2018Large}.
We train all the models for a maximum of 500 epochs using Adam optimizer \cite{kingma2014adam} to minimize the cross-entropy loss function, and the empirical learning rate is set as 0.003.

\begin{table}
	\renewcommand\arraystretch{1.5}
	\caption{Comparison with existing methods and baselines on validation set of ScanNet.}
	\centering
	\begin{tabular}{c|c|c}
		\hline
		Method        & Supervision & mIoU \\ \hline
		PointNet++ \cite{qi2017pointnet++}   & Point& 33.9 \\
		PointCNN \cite{li2018pointcnn}     &     Point        & 45.8 \\
		KPConv \cite{thomas2019kpconv}       &        Point     &   68.4   \\
		SparseConvNet \cite{choy20194d}       &        Point     &   73.6   \\
		SubsparseCNN \cite{graham20183d} &     Point        &  72.5    \\ \hline
		PCAM \cite{wei2020multi}       &     Scene        &  20.7    \\
		MPRM \cite{wei2020multi}         &       Scene      &     22.9 \\
		WyPR \cite{ren20213d}         &      Scene       &    29.6  \\
		WHCN      &      Scene       &   \textbf{37.3}  \\ \hline
		
	\end{tabular}
	\label{tab:result}
\end{table}

\begin{table*}[]
	\renewcommand\arraystretch{1.5}
	\caption{ The category-specific mIoU(\%) evaluation of pseudo labels on ScanNet training set.}
	\centering
	\resizebox{\textwidth}{!}{
		
		\begin{tabular}{c|cccccccccccccccccccc|c}
			\hline
			Method & wall & floor & cabinet & bed  & chair & sofa & table & door & window & bookshelf & picture & counter & desk & curtain  & fridge & shower & toilet & sink & bathtub & other & mIoU \\ \hline
			PCAM\cite{wei2020multi}    & 54.9 & 48.3  & 14.1    & 34.7 & 32.9  & \textbf{45.3} & 26.1  & 0.6  & 3.3    & \textbf{46.5} & 0.6     & 6.0  & 7.4 &26.9  & 0.0    & 6.1    & 22.3   & 8.2  & 52.0    & 6.1   & 22.1 \\
			MPRM\cite{wei2020multi}    & 47.3 & 41.1  & 10.4    & 43.2 & 25.2  & 43.1 & 21.5  & 9.8  & 12.3   & 45.0 & 9.0     & 13.9  & 21.1 & 40.9  & 1.8    & 29.4   & 14.3   & \textbf{27.4 }& \textbf{59.0}    & 10.0  & 24.4 \\
			Ours   &   \textbf{67.1}   &    \textbf{71.6}   &     \textbf{24.4}    &   \textbf{55.4}   &   \textbf{33.2}    &   42.0   &    \textbf{36.7}   &    \textbf{36.4}  &     \textbf{20.1}   &     43.5 &    \textbf{25.4 }    &   \textbf{16.8}   & \textbf{38.5}   &   \textbf{65.4} &     \textbf{26.1}   &    \textbf{43.9 }   &   \textbf{41.8}     &  24.1    &     48.8    &    \textbf{20.5}   &   \textbf{38.9}   \\ \hline
			
	\end{tabular}}
	\label{tab:pseudo1}
\end{table*}

\subsection{Evaluations on ScanNet}

In this section, we compare our proposed method with the state-of-the-art approaches that leverage scene-level labels on ScanNet dataset. PCAM \cite{wei2020multi} is a point class activation map with a KPConv \cite{thomas2019kpconv} backbone, and MPRM \cite{wei2020multi} adds multiple additional self-attention modules to PCAM. WyPR \cite{ren20213d} is a joint 3D segmentation and detection model using scene-level supervision. Besides, we also report benchmark results with fully supervised baselines.

Firstly, we present the qualitative segmentation results on selected rooms from the validation dataset in Fig. \ref{fig:resultScanNet}. From top to bottom, we sequentially visualize the RGB view, ground truth, and our final segmentation results.
Then, the quantitative results are displayed in Table \ref{tab:result}.
We can observe that, compared to the other weakly supervision approaches, our approach gives a new state-of-the-art performance with scene-level annotations. Specifically, our approach outperforms PCAM \cite{wei2020multi}, MPRM \cite{wei2020multi} approximately 80.2\% and 62.9\%, respectively. It can also be found that compared to WyPR \cite{ren20213d}, which is the state-of-the-art approach on this task, our method significantly gains an improvement of 26.0\% on mIoU. Another interesting observation is that our approach even achieves better performance than the fully supervised method PointNet++ \cite{qi2017pointnet++}. These results support the argument that WHCN can discover more objects in the non-salient region.

To evaluate the generated pseudo labels, we simply present the category-specific segmentation results of our pseudo labels on the training set in Table \ref{tab:pseudo1}.
As can be seen, our approach achieves better results than PCAM \cite{wei2020multi} and MPRM \cite{wei2020multi} with scene-level supervision. 
Specifically, we observe that our method can efficiently locate the dominant categories like wall, floor, and ceiling.
Besides, WHCN learns to segment more small object regions from the salient categories. Notably, pictures, windows, and fridges can be correctly classified at a higher rate than other approaches. 
The performance proves that our method can overcome the point imbalance among different categories to produce effective pseudo labels.

\begin{table}
	\renewcommand\arraystretch{1.5}
	\caption{Comparison with existing methods and baselines on validation set of S3DIS.}
	\centering
	\begin{tabular}{c|c|c}
		\hline
		Method        & Supervision & mIoU \\ \hline
		PointNet \cite{qi2017pointnet}   & Point & 41.1 \\
		PointCNN \cite{li2018pointcnn}     &     Point        & 57.3 \\
		TangentConv \cite{tatarchenko2018tangent}     &     Point        & 52.8 \\
		SegCloud \cite{tchapmi2017segcloud}       &        Point     &   48.9   \\
		SparseConvNet \cite{choy20194d}       &        Point     &   65.4   \\
		SuperpointGraph \cite{2018Large} &     Point        &  58.0    \\ 
		3D RNN \cite{ye20183d}       &     Point        &  53.4    \\
		Virtual MV-Fusion \cite{kundu2020virtual}        &       Point      &     65.4 \\ \hline 
		MPRM \cite{wei2020multi}         &       Scene      &     29.1 \\
		WyPR \cite{ren20213d}         &      Scene       &    22.3  \\
		WHCN      &      Scene       &   \textbf{39.6}  \\ \hline
		
	\end{tabular}
	\label{tab:S3DIS}
\end{table}

\begin{table*}[]
	\renewcommand\arraystretch{1.5}
	\caption{ The category-specific mIoU(\%) evaluation of pseudo labels on S3DIS training set.}
	\centering
	\resizebox{\textwidth}{!}{
		\begin{tabular}{c|ccccccccccccc|c}
			\hline
			Method   & ceiling & floor & wall & beam & column & window & door & chair & table & bookcase & sofa & board & clutter & mIoU \\ \hline
			MPRM \cite{wei2020multi}	&    64.2     &   71.4    &  50.7    &  0.0    &   \textbf{7.5 }    &    14.7    &  20.5    &  31.4     &    42.1   &    \textbf{35.2}      &   38.6   &   8.3    &     12.0    &   30.5   \\
			WHCN &  \textbf{82.2} &  \textbf{88.5}    &  \textbf{69.7}     &  \textbf{0.1 }   &  6.9    &   \textbf{26.6}    &   \textbf{39.1}     &   \textbf{47.1}   &  \textbf{50.1}     &   32.9    &       \textbf{39.5}         &   \textbf{18.1}    &    \textbf{35.2}     &   \textbf{41.2 }  \\ \hline
	\end{tabular}}
	\label{tab:pseudoS3DIS}
\end{table*}

\subsection{Evaluations on S3DIS}

We also evaluate our WHCN on the S3DIS dataset. Following previous works \cite{2018Large,xu2020weakly,zhang2021perturbed}, we use the fifth fold (Area-5) for evaluation.

\begin{figure}[!t]
	\centering
	\includegraphics[width=0.45\textwidth]{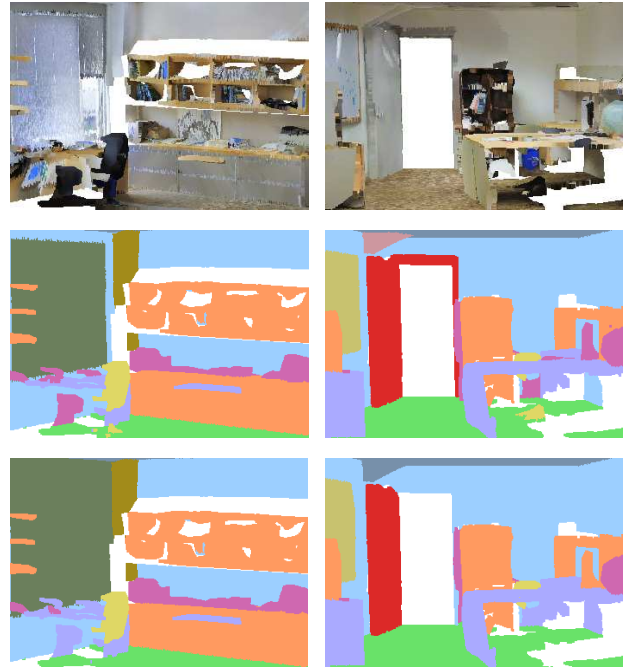}\\
	\caption{Qualitative results on the S3DIS dataset of our WHCN. The top row is the original input point clouds. The middle row is the ground truth. The bottom row shows the segmentation results with WHCN. } 
	\label{fig:resultS3DIS}
\end{figure}

In Table \ref{tab:S3DIS}, we provide the quantitative results of our algorithm. 
Similarly, we also compare our method with fully supervised approaches and other state-of-the-art weakly supervised approaches on Area-5 of S3DIS. 
It is observed that WHCN achieves a great improvement on average for scene-level supervision. 
With similar settings, WHCN increases by about 36.1\% and 77.6\% in terms of mIoU compared to the MPRM \cite{wei2020multi} and WyPR \cite{ren20213d}, respectively. 
The qualitative segmentation results on S3DIS are shown in Fig. \ref{fig:resultS3DIS}. 
It is observed that WHCN achieves good segmentation results.

Table \ref{tab:pseudoS3DIS} tabulates the category-specific evaluation of pseudo labels on the S3DIS dataset. 
It can also be found that WHCN can segment the objects from dominant categories well.
Meanwhile, compared to MPRM \cite{wei2020multi}, which is the state-of-the-art method in this task, our method gets promising results across object categories.
Specifically, WHCN gains 50.0\%, 90.7\% and 80.9\% improvements in categories chair, door, and window against MPRM \cite{wei2020multi}, respectively, which also indicates that our method can be well adapted to the point imbalance issue among different categories.

\subsection {Ablation Studies}

In this part, we further explore the contribution of the components of the proposed method for weakly supervised point cloud semantic segmentation. Then, we analyze the properties of hypergraph convolution with a series of ablation studies. All experiments are performed on ScanNet \cite{2017ScanNet} and S3DIS \cite{armeni2017joint}. Experiment \#1 only adopts baseline CAM method. \#2 and \#3 add the superpoint module and WHCN module, respectively. \#4 conducts the WHCN with an identity matrix for meaning equal weights for all hyperedges. \#5 represents the whole framework proposed in this paper.

As shown in Tabel \ref{tab:Ablation}, from the comparison of \#3 and \#4, we study the effectiveness of superpoint module.
It can be seen that compared to WHCN which only adopts the baseline CAM method (Base), the superpoint module (SP) achieves improvements of 11.3\% and 5.7\% in terms of mIoU on the ScanNet and S3DIS, respectively. The results show that the superpoint module can further improve the performances of the weakly supervised semantic segmentation. 

From the comparison of \#2 and \#4, we study the effectiveness of WHCN.
It can be found that if we only use the seed labels generated with baseline, the mIoU on the ScanNet and S3DIS are 26.3 and 26.8, respectively. With WHCN, the performance is significantly improved by 46.0\% and 51.1\%, indicating the effectiveness of our proposed label propagation mechanism. It enables an efficient information propagation between vertices by fully exploiting the high-order relationship.

From the comparison of \#4 and \#5, we study the effectiveness of hyperedge attention module. The results demonstrate that the hyperedge weighting scheme	can improve mIoU by 1.3\% and 1.7\% on the ScanNet and S3DIS, respectively, since the weighted hyperedges can better provide the intrinsic relationships between vertices.

\begin{table}[]
	
	\renewcommand\arraystretch{1.5}
	\caption{ The effectiveness of defferent components on ScanNet and S3DIS.}
	\centering
	\begin{tabular}{c|cccc|cc}
		\hline
		& Base & SP & Attention &WHCN & ScanNet & S3DIS \\ \hline
		\#1	& \checkmark     &         &  &     &    23.5  &  25.6    \\
		\#2	& \checkmark     & \checkmark        &     &   &  26.3    &  26.8    \\
		\#3	& \checkmark     &         &     &\checkmark  &  34.5    &   38.3   \\
		\#4	& \checkmark     & \checkmark     &   &  \checkmark     &   {38.4}   &  {40.5}    \\ 
		\#5	& \checkmark     & \checkmark        &  \checkmark   & \checkmark  &  \textbf{38.9}    &  \textbf{41.2}    \\ \hline
	\end{tabular}
	\label{tab:Ablation}
\end{table}

Finally, the comparison with baseline methods is given in Table \ref{tab:baseline}. The comparison is primarily done with DGCNN \cite{wang2019dynamic} and SPG \cite{2018Large}, which are two representatives of graph neural networks that have close relationships with our methods. The performance improvements brought by WHCN are 13.4\% and 11.6\% over DGCNN on the ScanNet and S3DIS datasets, respectively. Compared to SPG, WHCN improves the mIoU by 19.7\% and 24.1\% on the ScanNet and S3DIS datasets, respectively. This indicates that the higher-order hypergraph model can better exploit non-pairwise relationships.  Meanwhile, it aggregates all vertex attributes related to a hyperedge, and uses them to update the attribute of the hyperedge.

\begin{table}
	\renewcommand\arraystretch{1.5}
	\caption{Comparison with existing methods and baselines on ScanNet and S3DIS.}
	\centering
	\begin{tabular}{c|c|c}
		\hline
		Method        & ScanNet & S3DIS \\ \hline
		DGCNN \cite{wang2019dynamic}   & 34.3 & 36.9 \\
		SPG \cite{2018Large}   & 32.5 & 33.2 \\
		
		WHCN      &      \textbf{38.9}       &   \textbf{41.2}  \\ \hline
		
	\end{tabular}
	\label{tab:baseline}
\end{table}

\section{Conclusion}\label{sec.conclusion}
In this paper, we propose a new framework for the scene-level supervised point cloud semantic segmentation task. 
Specifically, we introduced a hypergraph-based label propagation unit to help the classification network capture global relations among the superpoints. 
This can strengthen the ability of the network to produce confident pseudo labels of objects.
We validate our method on two open datasets, and make extensive comparisons with the state-of-the-art approaches on this problem. 
The experimental results show that WHCN can achieve superior performance compared to other weakly supervised methods. 
Meanwhile, experiments also verify the effectiveness of our method regarding the point imbalance issue among different categories.
Although in this paper we only implement WHCN to generate pseudo labels from scene-level annotations, WHCN can also be applied to other annotations with appropriate transformations on the data sets. We intend to investigate these issues in future works.


%

\section*{Acknowledgment}

The authors sincerely acknowledge the anonymous reviewers for their insights and comments to further improve the quality of the manuscript.
They also would like to thank the participants in the study for their valuable time, and to the members of the Visual Computing Group at NJUST for their feedback. 
This work was supported by National Key R\&D Program of China (2018YFB1004904).

\ifCLASSOPTIONcaptionsoff
  \newpage
\fi



%
\bibliographystyle{IEEEtran}
\bibliography{weakly}

%

\begin{IEEEbiography}[{\includegraphics[width=1in,height=1.25in,clip,keepaspectratio]{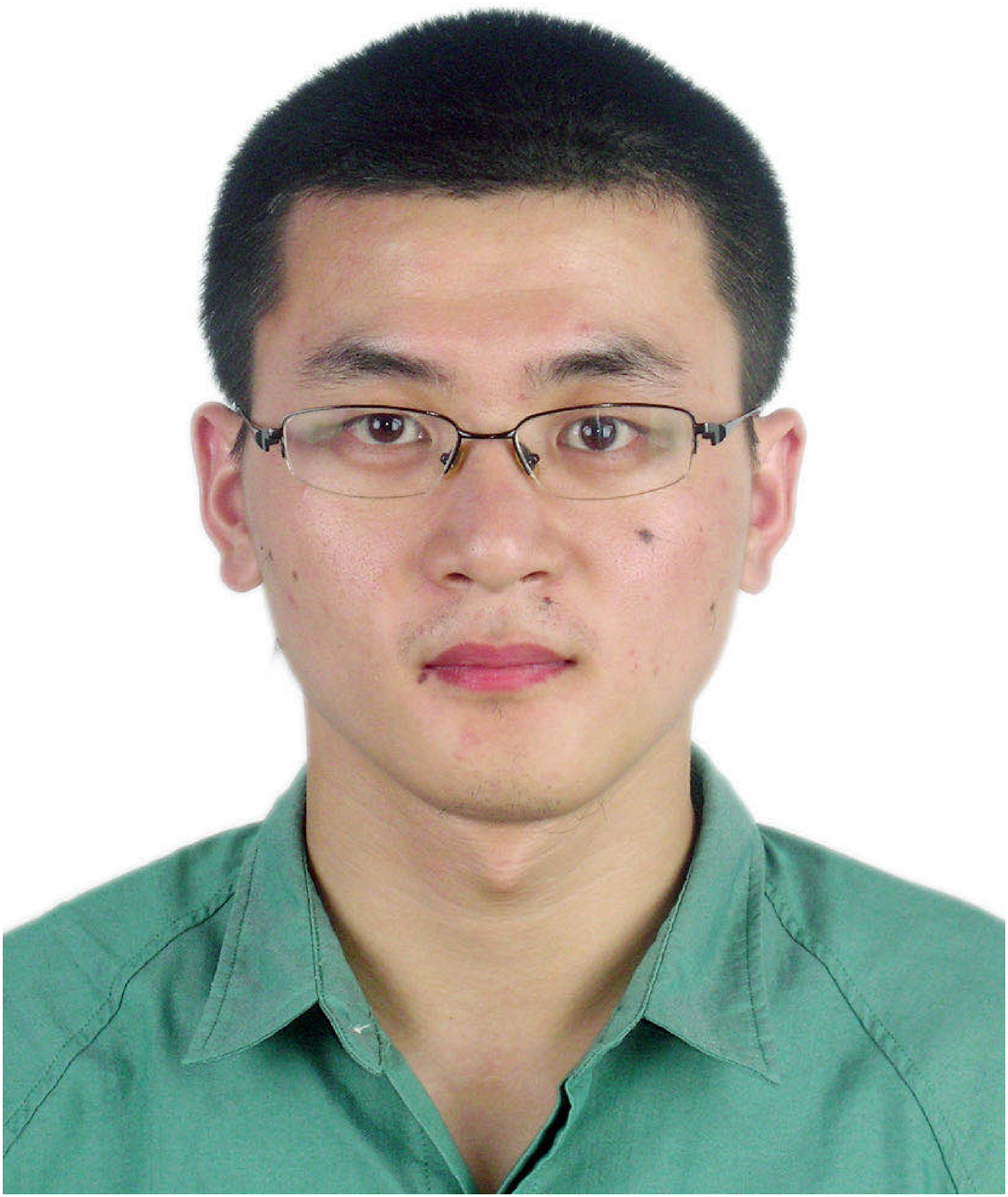}}]{Zhuheng Lu}
	
	is currently working toward the Ph.D. degree at the School of Automation, Nanjing University of Science and Technology, China. He received the M.S. degree from the School of Automation, Nanjing University of Science and Technology, in 2014. His research interests include computer graphics, 3D model recognition and machine learning.
\end{IEEEbiography}

\begin{IEEEbiography}[{\includegraphics[width=1in,height=1.25in,clip,keepaspectratio]{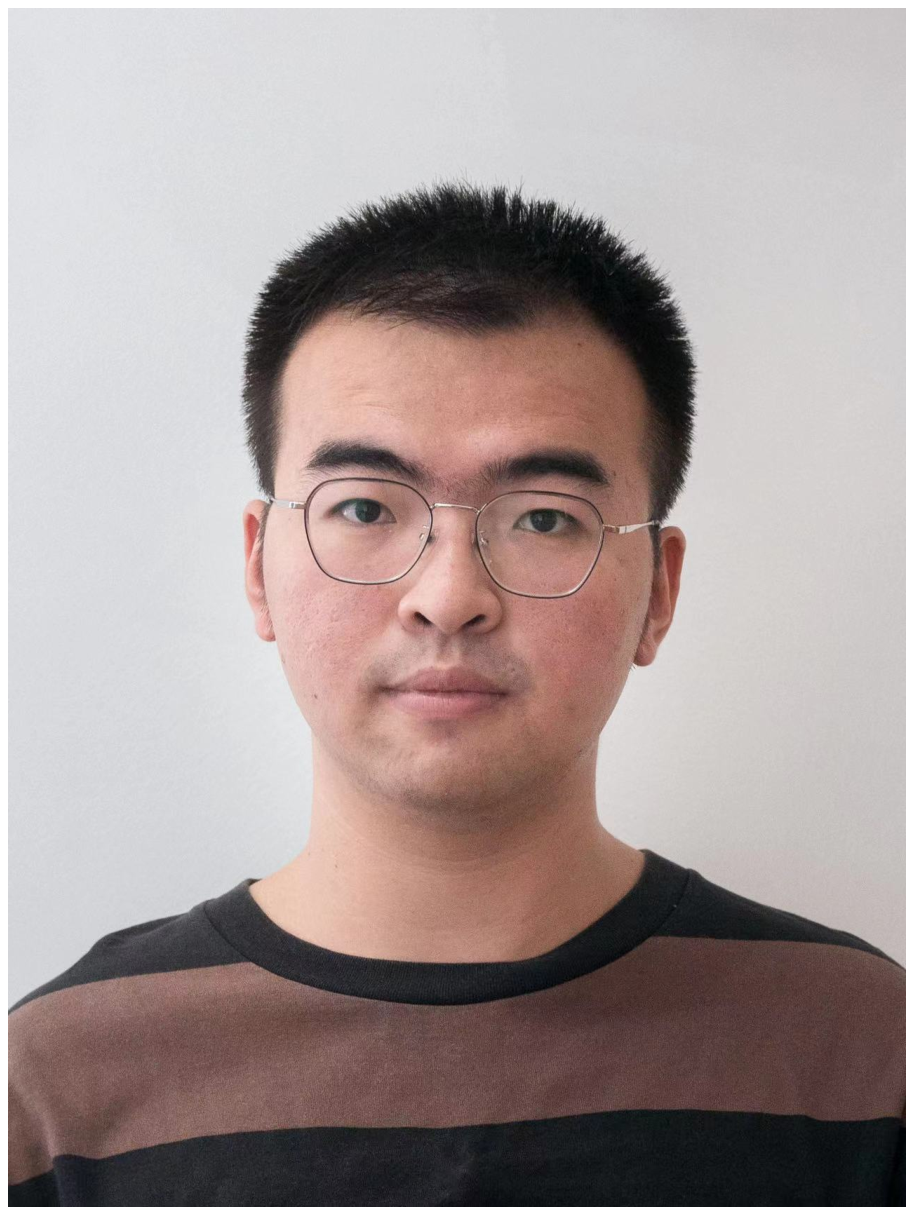}}]{Peng Zhang}
	is currently pursuing the M.S. degree at the School of Automation, Nanjing University of Science and Technology, China. He received the B.S. degree from Henan University of Technology, China, in 2020. His research interests include machine learning and 3d computer vision.
\end{IEEEbiography}

\begin{IEEEbiography}[{\includegraphics[width=1in,height=1.25in,clip,keepaspectratio]{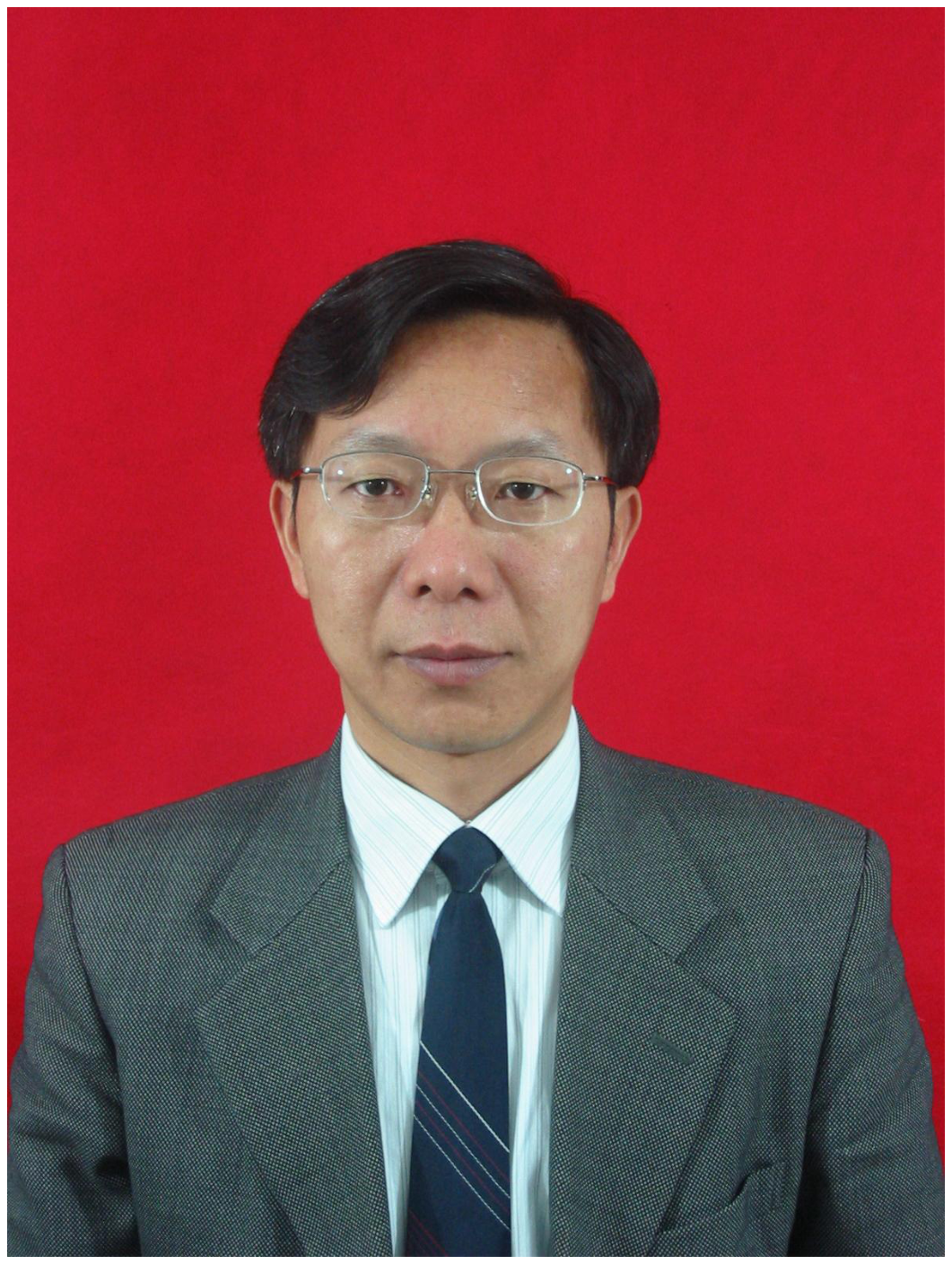}}]{Yuewei Dai}
	is currently a Professor at the School of Electronic and Information Engineering, Nanjing University of Information Science and Technology, China. He received the M.S. and Ph.D. degrees from the Nanjing University of Science and Technology, in 1987 and 2002, respectively, all in automation. His research interests include information security, signal, and image processing.
\end{IEEEbiography}
\begin{IEEEbiography}[{\includegraphics[width=1in,height=1.25in,clip,keepaspectratio]{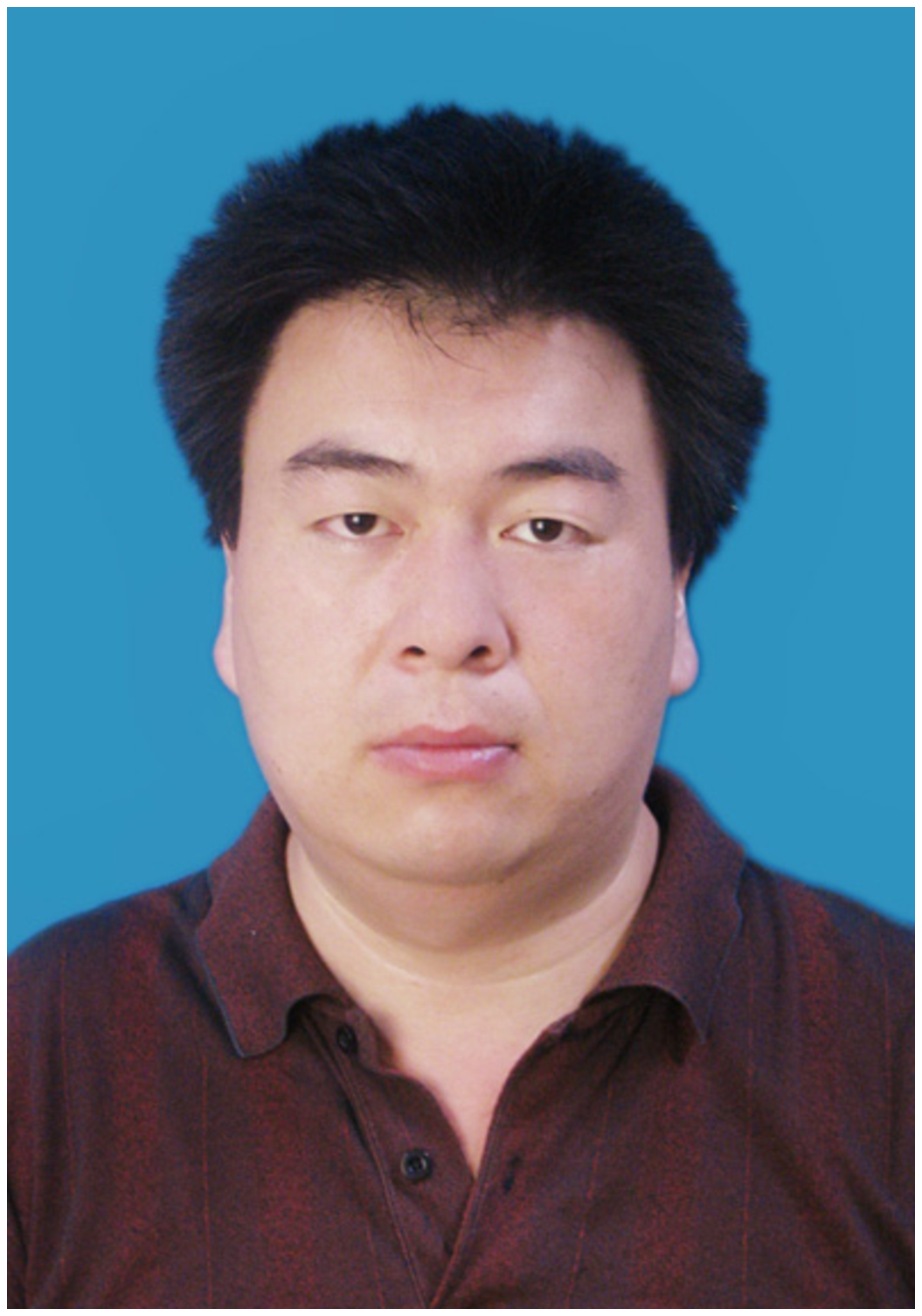}}]{Weiqing Li}
	is currently an associate professor at the School of Computer Science and Engineering, Nanjing University of Science and Technology, China. He received the B.S. and Ph.D. degrees from the School of Computer Sciences and Engineering, Nanjing University of Science and Technology in 1997 and 2007, respectively. His current interests include computer graphics and virtual reality.
\end{IEEEbiography}
\begin{IEEEbiography}[{\includegraphics[width=1in,height=1.25in,clip,keepaspectratio]{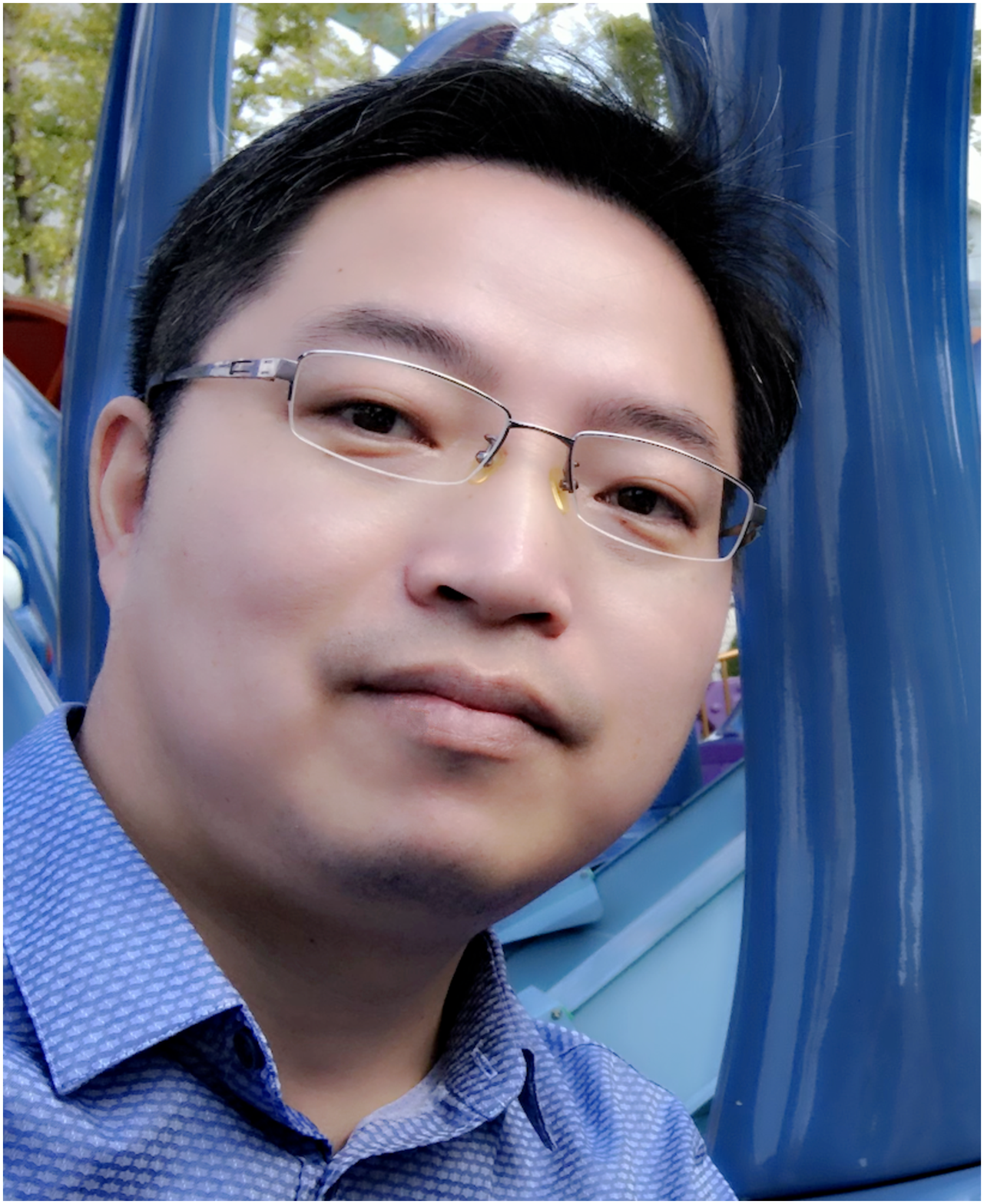}}]{Zhiyong Su}
	is currently an associate professor at the School of Automation, Nanjing University of Science and Technology, China. He received the B.S. and M.S. degrees from the School of Computer Science and Technology, Nanjing University of Science and Technology in 2004 and 2006, respectively, and received the Ph.D. from the Institute of Computing Technology, Chinese Academy of Sciences in 2009. His current interests include computer graphics, computer vision, augmented reality, and machine learning.  
\end{IEEEbiography}




\end{document}